%% file: main.tex
\documentclass[10pt,twocolumn,letterpaper]{article}

\usepackage[pagenumbers]{cvpr} 

\input{preamble}
\definecolor{cvprblue}{rgb}{0.21,0.49,0.74}
\usepackage[pagebackref,breaklinks,colorlinks,allcolors=cvprblue]{hyperref}
\usepackage{tikz,multirow,siunitx,booktabs}
\usepackage{bbding}
\usepackage[linesnumbered,ruled,vlined]{algorithm2e}

\def\paperID{12923} 
\def\confName{CVPR}
\def\confYear{2026}

\title{Decoupling Complexity from Scale in Latent Diffusion Model}

\author{%
Tianxiong Zhong, Xingye Tian, Xuebo Wang\footnotemark[2] , Boyuan Jiang, Xin Tao, Pengfei Wan \\
Kling Team, Kuaishou Technology \\
{\tt\small inkosizhong@gmail.com,} {\tt\small \{tianxingye,jiangboyuan,wangxuebo\}@kuaishou.com,} \\
{\tt\small \{taoxin,wanpengfei\}@kuaishou.com, \footnotemark[2]~Corresponding Author}
}

\begin{document}
\twocolumn[{%
\renewcommand\twocolumn[1][]{#1}%
\maketitle
\iftoggle{cvprfinal}{
    \vspace{-0.6cm}
}{}
\begin{center}
\captionsetup{type=figure}
\centering

\begin{subfigure}[b]{0.19\textwidth}
    \centering
    \includegraphics[width=\textwidth]{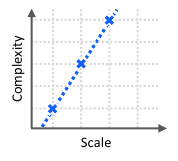}
    \caption{Vanilla LDM}
    \label{fig:assum_vanilla}
\end{subfigure}
\hfill
\begin{subfigure}[b]{0.19\textwidth}
    \centering
    \includegraphics[width=\textwidth]{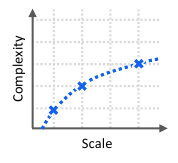}
    \caption{DetailFlow}
    \label{fig:assum_detailflow}
\end{subfigure}
\hfill
\begin{subfigure}[b]{0.19\textwidth}
    \centering
    \includegraphics[width=\textwidth]{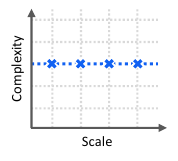}
    \caption{VFRTok}
    \label{fig:assum_vfrtok}
\end{subfigure}
\hfill
\begin{subfigure}[b]{0.19\textwidth}
    \centering
    \includegraphics[width=\textwidth]{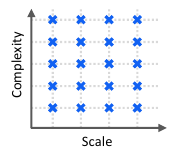}
    \caption{\textbf{Ours}}
    \label{fig:assum_ours}
\end{subfigure}
\hfill
\begin{subfigure}[b]{0.19\textwidth}
    \centering
    \includegraphics[width=\textwidth]{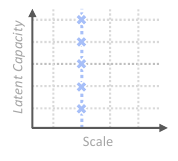}
    \caption{FlexTok}
    \label{fig:assum_flextok}
\end{subfigure}

\caption{
Comparison of how different latent diffusion models handle the relationship between content complexity and scale.
(a–b) Couple scale and complexity, requiring larger latents for higher resolutions or frame rates.
(c) Models only a fixed level of complexity.
\textbf{(d) Our approach decouples complexity from scale, enabling flexible and efficient trade-offs.}
(e) Utilizes arbitrary length latents to model fixed level of both scale and complexity based on a flow-based decoder.
}
\label{fig:assum}
\end{center}%
}]


\input{sec/0_abstract}    

\input{sec/1_intro}
\input{sec/2_relate}

\input{sec/3_method}
\input{sec/4_exp}
\input{sec/5_conclusion}

{
    \small
    \bibliographystyle{ieeenat_fullname}
    \bibliography{main}
}

\input{sec/X_suppl}

\end{document}

%% file: sec/0_abstract.tex
\begin{abstract}
Existing latent diffusion models typically couple scale with content complexity, using more latent tokens to represent higher-resolution images or higher–frame rate videos.
However, the latent capacity required to represent visual data primarily depends on content complexity, with scale serving only as an upper bound.
Motivated by this observation, we propose DCS-LDM, a novel paradigm for visual generation that decouples information complexity from scale.
DCS-LDM constructs a hierarchical, scale-independent latent space that models sample complexity through multi-level tokens and supports decoding to arbitrary resolutions and frame rates within a fixed latent representation.
This latent space enables DCS-LDM to achieve a flexible computation–quality tradeoff.
Furthermore, by decomposing structural and detailed information across levels, DCS-LDM supports a progressive coarse-to-fine generation paradigm.
Experimental results show that DCS-LDM delivers performance comparable to state-of-the-art methods while offering flexible generation across diverse scales and visual qualities.
\end{abstract}

%% file: sec/1_intro.tex
\section{Introduction}
Latent Diffusion Models (LDMs) have recently emerged as a powerful framework for visual generation~\cite{rombach2022ldm,li2024hunyuan,wan2025wan,yang2024cogvideox,ma2025stept2v,agarwal2025cosmos}.
An LDM employs a tokenizer to embed high-dimensional images or videos into a compressed latent representation~\cite{chen2024dcae,chen2025dcae1.5,chen2025maetok,zhong2025vfrtok,wang2024omnitokenizer}, followed by a diffusion model that performs denoising in the latent space~\cite{peebles2023dit,ma2024sit,yao2025lightningdit,yu2024repa}.
This latent space inherently provides a rate–distortion trade-off~\cite{berger2003rd,lu2022learning}: larger latent representations possess higher capacity to preserve fine details of the visual data, but at the cost of increased computational overhead.
Meanwhile, users exhibit diverse demands for generation scale, such as varying resolutions and frame rates.
For example, mobile users may prefer high–frame rate, vertical videos optimized for smartphone displays, whereas professional creators may seek ultra–high resolution, horizontal videos tailored for large cinema screens.

As illustrated in \Cref{fig:assum}, mainstream LDMs couple the scale of visual data with its content complexity.
They allocate more latent tokens to higher-resolution or higher–frame rate data, assuming that scale directly reflects the complexity.
However, this assumption overlooks the variability in information content across samples.
While input size determines the number of pixels, the upper bound of representable information, the actual information content depends primarily on the semantic and structural complexity of the visual data.
For example, a large-scale animation may contain far less information than a small-scale nature documentary.
From an information–theoretic perspective, this implies that each sample should have its own rate–distortion curve, meaning that the number of latent tokens required to achieve comparable fidelity should vary across samples.

\begin{figure}
    \centering
    \begin{subfigure}[b]{\linewidth}
        \centering
        \includegraphics[width=\linewidth]{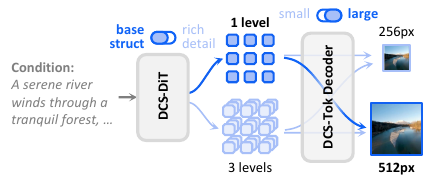}
        \caption{DCS-LDM decouples information complexity from data scale. The latents generated by DCS-DiT are hierarchical, capturing information ranging from global structure to fine details. The DCS-Tok decoder can decode data from these latents at arbitrary scales.}
        \label{fig:overview_decoupling}
    \end{subfigure} \\
    \begin{subfigure}[b]{\linewidth}
        \centering
        \includegraphics[width=\linewidth]{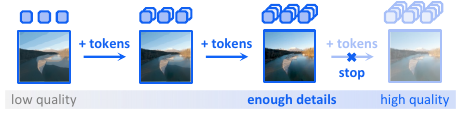}
        \caption{DCS-LDM supports progressive generation from coarse to fine. Users can incrementally increase the number of generated tokens and flexibly decide when to stop based on the quality of the results.}
        \label{fig:overview_coarse_to_fine}
    \end{subfigure}
    \caption{Overview of DCS-LDM.}
    \label{fig:overview}
\end{figure}

Motivated by these insights, we propose DCS-LDM, a \textbf{L}atent \textbf{D}iffusion \textbf{M}odel that \textbf{D}ecouples information \textbf{C}omplexity from the sample \textbf{S}cale.
As illustrated in \Cref{fig:assum_ours}, DCS-LDM decomposes complexity and scale into two orthogonal dimensions, enabling flexible representation of diverse samples.
Specifically, it employs a scale-independent latent space, aligning images and videos of different scales into a unified representation, and structures this space hierarchically, where additional token levels model higher information complexity.
As shown in \Cref{fig:overview}, DCS-LDM comprises two key components: the tokenizer DCS-Tok and the transformer-based diffusion model DCS-DiT.
During generation, quality is controlled by the number of latent token levels generated by DCS-DiT, while scale can be freely chosen during DCS-Tok decoding.

These designs enable DCS-LDM to achieve a flexible computation–quality tradeoff.
We further structure the latent space causally, assigning basic structural information to earlier levels and fine-grained details to later ones.
DCS-LDM thus supports a coarse-to-fine progressive generation paradigm, maintaining identical computational cost while reducing latency and enabling efficient refinement~\cite{hu2024vivid}.

In summary, our main contributions are as follows:
\begin{enumerate}
\item We reveal the inherent differences in content complexity among visual samples and introduce a decoupling perspective between scale and complexity.
\item We propose DCS-LDM, a novel latent diffusion paradigm that naturally supports flexible computation–quality trade-offs and coarse-to-fine generation.
\item Experiments demonstrate that DCS-LDM achieves comparable reconstruction and generation performance to state-of-the-art methods, while supporting more diverse and scalable functionalities.
\end{enumerate}

%% file: sec/2_relate.tex
\begin{figure*}[t]
    \centering
    \includegraphics[width=\textwidth]{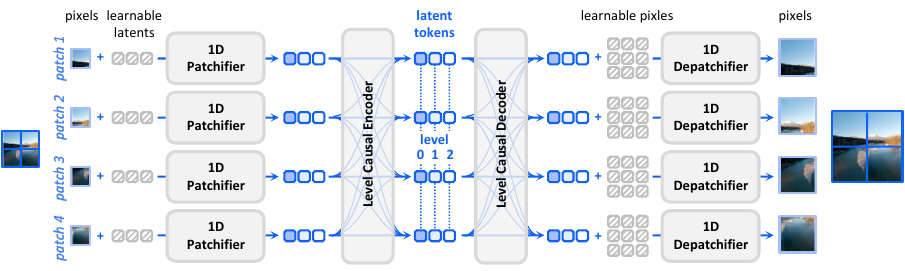}
    \caption{DCS-Tok adopts a pure Transformer architecture, comprising a 1D patchifier–depatchifier and a causal encoder–decoder. The patchifier–depatchifier align scale-aware spatial–temporal patches into a causal token sequence, where lower-level tokens capture structure and higher-level tokens encode fine-grained textures. The encoder–decoder facilitate cross-patch interaction while preserving causal dependencies across token levels.}
    \label{fig:arch_tok}
\end{figure*}

\section{Related Work}
\subsection{Vanilla Latent Diffusion Models}
LDMs~\cite{rombach2022ldm,peebles2023dit,ma2024sit,yao2025lightningdit,yu2024repa} were introduced to improve the efficiency of pixel-space diffusion models~\cite{ho2020ddpm,song2020ddim}.
An LDM typically consists of two core components: a visual tokenizer and a diffusion model.
This design has been widely adopted in both image~\cite{yao2025lightningdit,chen2024softvq,chen2025maetok,rombach2022ldm,bachmann2025flextok,liu2025detailflow,li2024hunyuan,sun2024hunyuanlarge,wan2025wan} and video~\cite{agarwal2025cosmos,hacohen2024ltx,ma2025stept2v,kong2024hunyuanvideo,yang2024cogvideox,zheng2024opensora,zhong2025vfrtok} generation, driving rapid progress in visual generative modeling.
In LDMs, the capacity of the latent space plays a crucial role in balancing efficiency and visual quality.
The tokenizer bridges the pixel space and the latent space, providing diverse strategies for information compression and representation.
Vanilla LDMs~\cite{rombach2022ldm,yao2025lightningdit,li2024hunyuan,yang2024cogvideox} commonly employ CNN-based tokenizers to perform fixed-ratio downsampling.
For example, in video generation, it is typical to downsample the spatial and temporal dimensions by factor of 16 and 4, respectively.
Consequently, as the data scale increases, the number of latent tokens expands linearly with scale (\Cref{fig:assum_vanilla}).

\subsection{Novel Visual Tokenizers}
Recently, several 1D tokenizers have been proposed~\cite{yu2024titok,chen2024softvq,chen2025maetok,zhong2025vfrtok,liu2025detailflow}, which map grid-based visual patches into a 1D sequence, removing explicit spatial relationships.
These methods typically adopt ViT-based architectures.
During encoding, the input image or video is divided into patches, learnable query tokens are concatenated, and only these tokens are retained as latents.
During decoding, the query tokens are concatenated to the latent tokens to reconstruct each patch of the original data.

This flexible formulation allows for the construction of tokenizers under different information assumptions, controlled by the number of patch or latent tokens~\cite{zhong2025vfrtok,bachmann2025flextok,liu2025detailflow}.
For instance, VFRTok~\cite{zhong2025vfrtok} aligns a varying number of patch tokens into a fixed-length latent, enabling variable frame-rate generation.
FlexTok~\cite{bachmann2025flextok} and SEMANTICIST~\cite{wen2025semanticist} instead model fixed-size images using latents of arbitrary length.
However, their flow-based decoding paradigm introduces randomness, meaning that increasing the number of token may not refine the generated results but entirely alter image content.
DetailFlow~\cite{liu2025detailflow} further explores the sub-linear relationship between scale and complexity from an entropy perspective.
However, the existing visual tokenizers overlook the variation in content complexity across samples, which limits their ability to adapt latent capacity.

%% file: sec/3_method.tex
\section{Method}
\begin{figure*}[t]
    \centering
    \includegraphics[width=0.9\textwidth]{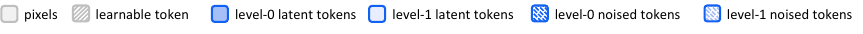} \\
    \vspace{-0.24cm}
    \begin{subfigure}[t]{0.43\textwidth}
        \centering
        \includegraphics[width=\textwidth]{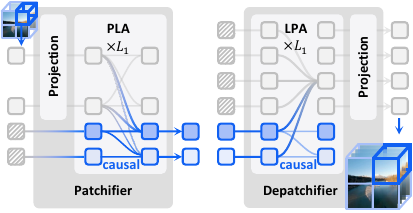}
        \caption{Each pixel is projected as an independent token. Lines indicate information flow: pixels to latents in the patchifier, latents to pixels in the depatchifier, and causal between latent tokens.}
        \label{fig:arch_patchify}
    \end{subfigure}
    \hfill
    \begin{subfigure}[t]{0.275\textwidth}
        \centering
        \includegraphics[width=\textwidth]{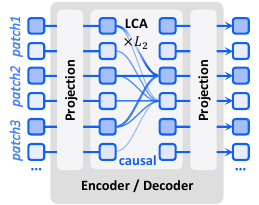}
        \caption{Lines indicate information flow: a token can attend only to tokens at its own level or lower across all patches.}
        \label{fig:arch_autoencoder}
    \end{subfigure}
    \hfill
    \begin{subfigure}[t]{0.245\textwidth}
        \centering
        \includegraphics[width=\textwidth]{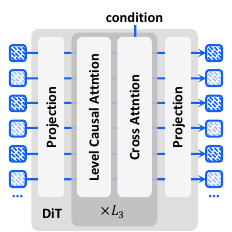}
        \caption{DCS-DiT extends (b) by adding cross attention blocks for conditional information injection.}
        \label{fig:arch_dit}
    \end{subfigure}
    
    \caption{Detailed architectures. (a) Patchifier and depatchifier of DCS-Tok. (b) Encoder and decoder of DCS-Tok. (c) DCS-DiT. PLA, LPA, and LCA denote Pixel-to-Latent, Latent-to-Pixel, and Level Causal Attention, respectively. $L_i$ indicates the number of blocks.}
    \label{fig:arch}
\end{figure*}

To construct the hierarchical, scale-independent latent space, we introduce a carefully designed tokenizer (\Cref{fig:arch_tok}) together with a corresponding DiT architecture (\Cref{fig:arch_dit}).
The tokenizer incorporates a novel 1D patchifier that operates on a fixed number of patches, enabling the model to encode inputs of varying scales into a unified multi-level latent representation.
More levels of latent tokens capture fine-grained details for complex samples, while fewer levels suffice for simpler ones, thereby effectively decoupling scale from complexity.

To better illustrate how the model structure aligns with our design principles, we detail each component according to its objective in the following sections.
We first describe how data of different scales are mapped into a unified latent space in \Cref{sec:latents_scale}.
Then, we present the construction of hierarchical representations in \Cref{sec:latents_level}.
Finally, we discuss the coarse-to-fine generation paradigm in detail in \Cref{sec:coarse_to_fine}.
Additional implementation details, such as positional embeddings, attention masks, and temporal causality, are provided in the supplementary material (\Cref{supp_sec:details}).

\subsection{Scale-independent Latent Space}\label{sec:latents_scale}
\noindent\textbf{Scale-aware Patchify.}
As detailed in \Cref{fig:arch_tok}, instead of fixing the patch size~\cite{rombach2022ldm,wang2024omnitokenizer,yao2025lightningdit,li2024hunyuan,wan2025wan}, we fix the number of patches, and determine the size based on its resolution and frame rate.
Specifically, for a video with resolution $h\times w$ and frame rate $f$, the spatial and temporal patch sizes, $p_{h,w}$ and $p_t$, are defined as:
\begin{equation}
p_{h,w}=\frac{\min(h,w)}{k},\quad p_t=\frac{f}{k_t},
\label{eq:patch_size}
\end{equation}
where $k$ is the number of patches along the shorter spatial dimension and $k_t$ is the number of patches within a second.

For example, as illustrated in \Cref{fig:arch_patchify}, the same video is patchified at different scales with $k=2$ and $k_t=1$.
The smaller one (top-left) has a resolution of $16\times16$ and frame rate 2 fps, resulting in $p_{h,w}=8$ and $p_t=2$.
The larger copy (bottom-right) has a resolution of $32\times32$ and frame rate 4 fps, giving $p_{h,w}=16$ and $p_t=4$.
This approach aligns the same spatial–temporal regions across scales.
For images, we treat them as single-frame videos with $p_t=1$.

\noindent\textbf{Asymmetric scale alignment.}
To enable decoding at arbitrary resolutions and frame rates from a unified latent representation, DCS-Tok aligns patches across different scales.
As shown in \Cref{fig:arch_tok}, we adopt asymmetric training~\cite{zhong2025vfrtok} where the input and output of DCS-Tok are randomly chosen scale variants of the same sample during training.

\subsection{Hierarchical Representation}\label{sec:latents_level}
\noindent\textbf{1D patchifier.}
As detailed in \Cref{fig:arch_patchify}, we implement the patchifier and depatchifier in a 1D tokenizer style~\cite{yu2024titok,chen2025maetok,zhong2025vfrtok}, using learnable tokens to query information from the pixels in each patch.
All pixels in a patch are treated as independent tokens and concatenated with $n$ learnable latent tokens, numbered sequentially from level-1 to level-$n$.
Symmetrically, in the depatchifier, learnable pixel tokens are concatenated before the latent tokens to reconstruct pixel information.
This architecture enables DCS-Tok to encode scale-aware patches into a hierarchical representation.

\noindent\textbf{Autoencoder.}
Since the patchifier processes each patch independently, the autoencoder is responsible for modeling inter-patch interactions.
As shown in \Cref{fig:arch_autoencoder}, the encoder and decoder share the same architecture.
The resulting latent representation preserves the spatio-temporal structure of the input while introducing an additional hierarchical dimension.
For a video with $t$ frames and spatial resolution $h \times w$, the representation has the shape
$\frac{t}{p_t} \times \frac{h}{p_{h,w}} \times \frac{w}{p_{h,w}} \times n$,
where $n$ denotes the number of levels.

\noindent\textbf{DiT.}
DCS-DiT models the transformation between the hierarchical latents and pure noise, with the number of generated levels determined by the user-specified noise level.
To support hierarchical latent token generation, DCS-DiT extends conventional DiTs~\cite{yao2025lightningdit,peebles2023dit,ma2024sit}.
As shown in \Cref{fig:arch_dit}, it adopts an architecture similar to the autoencoder in DCS-Tok and introduces optional cross-attention blocks for injecting conditional information.

\subsection{Coarse-to-fine Generation}\label{sec:coarse_to_fine}
\noindent\textbf{Level causality.}
All components of DCS-LDM are designed with level causality, ensuring that each latent level depends only on the preceding one.
As shown in \Cref{fig:arch}, we illustrate three key mechanisms: Pixel-to-Latent Attention (PLA), Latent-to-Pixel Attention (LPA), and Level Causal Attention (LCA).
We also highlight the information flow within different modules.
Unlike prior 1D tokenizer designs~\cite{yu2024titok,chen2025maetok,chen2024softvq,zhong2025vfrtok} that use full attention across all tokens, PLA and LPA compute full attention only among pixel tokens.
Information exchange between pixel and latent tokens is strictly unidirectional, preventing violation of level causality among latent tokens.
For LCA, information is allowed to flow across patches, but strictly in a unidirectional manner: information from lower levels can influence higher levels, but not vice versa.

\noindent\textbf{Structured representation.}
To enable reconstruction from an arbitrary number of token levels, our objective is to encode essential structural information in the early levels and reserve later levels for fine-grained details.
To achieve this, we randomly assign each patch between $1$ and $n$ token levels during training.
Because DCS-Tok is optimized to reconstruct patches even when higher-level tokens are missing, it naturally learns the desired coarse-to-fine encoding scheme.
Operationally, when a patch is assigned $1 \leq m \leq n$ token levels, an attention mask is applied to deactivate the remaining $n - m$ tail-level tokens.
During inference (batch size = 1), this simply amounts to discarding the unused tokens.
More details are provided in \Cref{supp_sec:attn_mask}.

\begin{figure}[t]
    \centering
    \includegraphics[width=\linewidth]{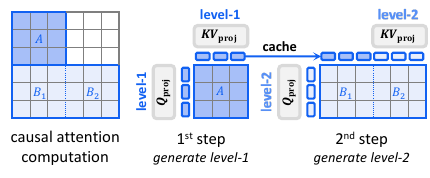}
    \caption{Coarse-to-fine generation. Causality enables computations to be separated by token levels. Caching keys and values from previous levels keeps the computational cost of progressive generation identical to that of non-progressive generation.}
    \label{fig:coarse_to_fine}
\end{figure}

\begin{figure*}[t]
    \centering
    \begin{subfigure}[t]{0.255\textwidth}
        \centering
        \includegraphics[width=\textwidth]{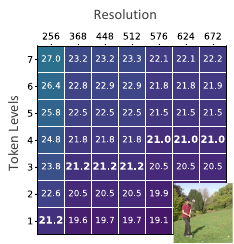}
        \caption{Complex content}
        \label{plot:complex1}
    \end{subfigure}
    \hfill
    \begin{subfigure}[t]{0.23\textwidth}
        \centering
        \includegraphics[width=\textwidth]{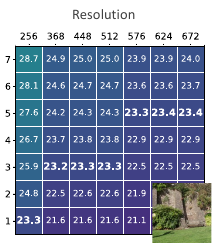}
        \caption{Complex content}
        \label{plot:complex2}
    \end{subfigure}
    \hfill
    \begin{subfigure}[t]{0.23\textwidth}
        \centering
        \includegraphics[width=\textwidth]{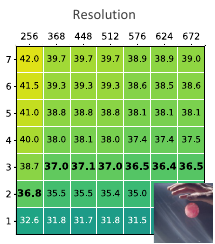}
        \caption{Simple content}
        \label{plot:simple1}
    \end{subfigure}
    \hfill
    \begin{subfigure}[t]{0.27\textwidth}
        \centering
        \includegraphics[width=\textwidth]{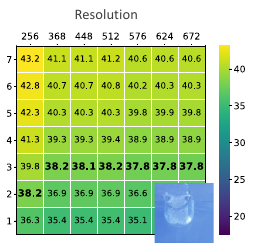}
        \caption{Simple content}
        \label{plot:simple2}
    \end{subfigure}
    \caption{Image reconstruction via DCS-Tok across samples with varying information complexity, resolution, and latent token levels. Brighter background indicates higher PSNR. Configurations with comparable PSNR are highlighted in bold.}
    \label{plot:complexity_recon_image}
\end{figure*}

\begin{figure}[t]
    \centering
    \includegraphics[width=\linewidth]{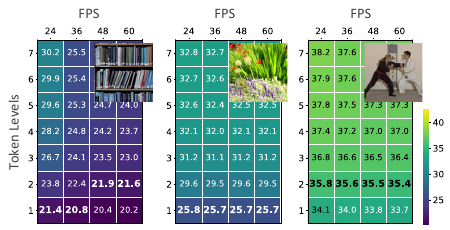}
    \caption{Video reconstruction via DCS-Tok with varying information complexity, frame rate, and latent token levels. Configurations with comparable PSNR are highlighted in bold.}
    \label{plot:complexity_recon_video}
\end{figure}

\noindent\textbf{Cache-based acceleration.}
With the above components in place, DCS-LDM naturally supports coarse-to-fine generation.
By introducing a caching mechanism, the model can switch from generating all $m$ token levels at once to generating them sequentially, while keeping the total computational cost unchanged.
Using two token levels as an example, \Cref{fig:coarse_to_fine} illustrates the attention operations within each DCS-DiT block and the DCS-Tok decoder.
Let $x_i$ denote all tokens at level-$i$.
The tokens in each level are first projected into query, key, and value features as:
\begin{align}
q_1 = Q_\mathrm{proj}(x_1),\,k_1 = K_\mathrm{proj}(x_1),\,v_1 = V_\mathrm{proj}(x_1),
\label{eq:proj1} \\
q_2 = Q_\mathrm{proj}(x_2),\,k_2 = K_\mathrm{proj}(x_2),\,v_2 = V_\mathrm{proj}(x_2).
\label{eq:proj2}
\end{align}
As highlighted in \Cref{fig:coarse_to_fine}, the effective attention computations include parts $A$, $B_1$, and $B_2$, defined as:
\begin{align}
&A:\mathrm{attn}(q_1, k_1, v_1),
\label{eq:attn1} \\
B_1:\mathrm{attn}&(q_2, k_1, v_1),\,B_2:\mathrm{attn}(q_2, k_2, v_2).
\label{eq:attn2}
\end{align}
During coarse-to-fine generation, level-1 tokens are first generated using \Cref{eq:proj1,eq:attn1}, followed by level-2 tokens using \Cref{eq:proj2,eq:attn2}.
Since the computation of $B_1$ depends on $k_1$ and $v_1$, these features are cached~\cite{dai2019transformerxl} from the first step, enabling efficient progressive generation.

\begin{figure}[t]
    \centering
    \includegraphics[width=\linewidth]{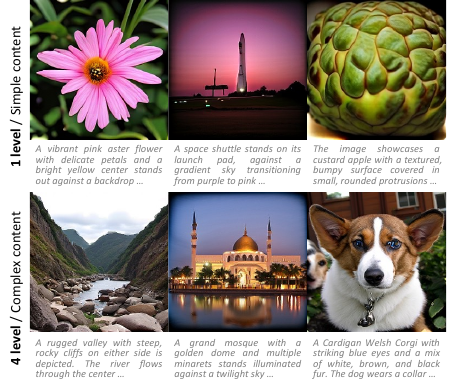}
    \caption{Text-to-image generation across samples with varying information complexity and latent token counts.}
    \label{plot:complexity_gen}
\end{figure}

%% file: sec/4_exp.tex
\begin{figure*}[t]
    \centering
    \begin{subfigure}[t]{\textwidth}
        \centering
        \includegraphics[width=\textwidth]{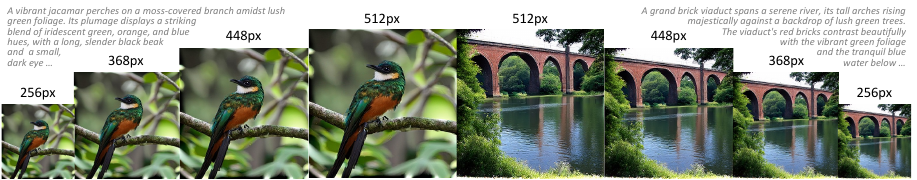}
        \caption{Text-to-image results at various resolutions using \textbf{4 levels} of token.}
        \label{plot:var_res}
    \end{subfigure}
    \\
    \begin{subfigure}[t]{0.52\textwidth}
        \centering
        \includegraphics[width=\textwidth]{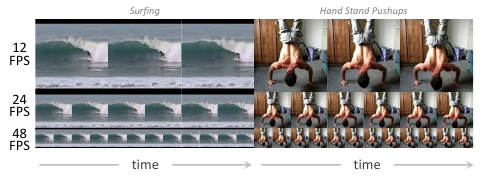}
        \caption{Label-to-video at various frame rates using \textbf{1 level} of token.}
        \label{plot:var_fps}
    \end{subfigure}
    \hfill
    \begin{subfigure}[t]{0.47\textwidth}
        \centering
        \includegraphics[width=\textwidth]{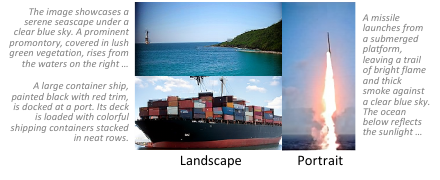}
        \caption{Text-to-image results at various aspect ratios.}
        \label{plot:var_ratio}
    \end{subfigure}
    \caption{DCS-LDM generates data at varying resolutions and frame rates via scale-independent latents and supports flexible aspect ratios.}
    \label{plot:var_scale}
\end{figure*}

\section{Experiments}
\subsection{Setup}
\noindent\textbf{Implementation.}
We employ ViT-B~\cite{dosovitskiy2020vit} as the backbone for the encoder and decoder ($L_2=12$), a tiny ViT ($L_1=3$) for the patchifier and depatchifier in DCS-Tok, and ViT-XL for DCS-DiT ($L_3=24$).
Following LightningDiT~\cite{yao2025lightningdit}, we adopt modern Transformer components such as SwiGLU FFN~\cite{shazeer2020swiglu} and RMSNorm~\cite{zhang2019rmsnorm}.
We set $k=16$, with $k_t=1$ for images and $k_t=6$ for videos, corresponding to a spatial downsampling factor of $p_{h,w}=16$ at $256\times256$ resolution and a temporal factor of $p_t=4$ at 24 fps~\cite{wan2025wan,yao2025lightningdit}.
Both label-guided and text-guided generation are implemented using Classifier-Free Guidance~\cite{ho2022cfg} (CFG), where the cross-attention blocks are dropped in the label-guided image and video generation.

We train DCS-Tok as an deterministic tokenizer, with reconstruction and latent regularization objectives.
\begin{equation}
\mathcal{L}_\mathrm{Tok}=\mathcal{L}_\text{recon}+\lambda_1\mathcal{L}_\text{percept}+\lambda_2\cdot\lambda_\nabla\mathcal{L}_\text{adv}+\lambda_3\mathcal{L}_\text{L2-margin},
\end{equation}
where $\mathcal{L}_\text{recon}$, $\mathcal{L}_\text{percept}$, $\mathcal{L}_\text{adv}$ are the L2 reconstruction loss, perceptual loss~\cite{larsen2016perceptual,johnson2016perceptual}, and adversarial loss~\cite{goodfellow2020gan}, respectively.
$\mathcal{L}_\mathrm{L2\text{-}margin}$ indicates L2 latent regularization with margin, which is detailed in \Cref{supp_sec:train}.
The factors are set to $\lambda_1=1$, $\lambda_2=0.2$, $\lambda_3=0.001$, and $\lambda_\nabla$ represents adaptive weight~\cite{chen2025maetok,zhong2025vfrtok}.
The entire training process is divided into multiple stages: low resolution image initialization, spatial asymmetric training, and spatial-temporal asymmetric training.
Furthermore, to improve the diffusibility of the high-level tokens, we additionally introduced the latent denoising technique~\cite{yang2025denoise} during the training.
DCS-DiT follows the sampling method and training objectives of LightiningDiT~\cite{yao2025lightningdit}.
All models are trained on four nodes, each equipped with eight GPUs.
Training DCS-Tok takes approximately one week, while DCS-DiT requires about four days.
More details are given in the \Cref{supp_sec:train}.

\noindent\textbf{Datesets.}
The training data for DCS-Tok comprise the ImageNet~\cite{deng2009imagenet}, Koala-36M~\cite{wang2025koala}, and LAVIB~\cite{stergiou2024lavib} datasets.
The Koala-36M and LAVIB dataset provides videos with relatively high resolutions and frame rates, allowing DCS-Tok to be trained across diverse configurations ranging from 128p to 720p and 12 fps to 60 fps.
For DCS-DiT, due to computational constraints, the qualitative experiments are conducted on a text-to-image model trained using the ImageNet~\cite{deng2009imagenet} and SA-1B~\cite{kirillov2023SAM} datasets, with captions generated by Qwen-VL 2.5~\cite{bai2025qwen2.5vl}.
Moreover, to ensure fair comparison with existing methods~\cite{chang2022maskgit,sun2024llamagen,bachmann2025flextok,liu2025detailflow,peebles2023dit,ma2024sit,yu2024repa,yao2025lightningdit,chen2025maetok,wang2024omnitokenizer,zhong2025vfrtok,agarwal2025cosmos}, we additionally train label-based image and video generation models.
For label-to-image generation, the ImageNet dataset~\cite{deng2009imagenet} is used for both training and evaluation, while for label-to-video generation, the UCF101 dataset~\cite{soomro2012ucf101} is incorporated for both training and evaluation.

\noindent\textbf{Metrics.}
We evaluate reconstruction quality using PSNR and SSIM for pixel-level assessment, and Fréchet Inception Distance (rFID)~\cite{heusel2017fid} and Fréchet Video Distance (rFVD)~\cite{unterthiner2019fvd} for perceptual quality on images and videos, respectively.
For generation, with and without Classifier-Free Guidance (CFG)~\cite{ho2022cfg}, we employ gFID~\cite{heusel2017fid} and gFVD~\cite{unterthiner2019fvd} to measure generation quality, and Inception Score (IS)~\cite{salimans2016IS} to assess semantic diversity.

\begin{figure*}[t]
    \centering
    \includegraphics[width=\linewidth]{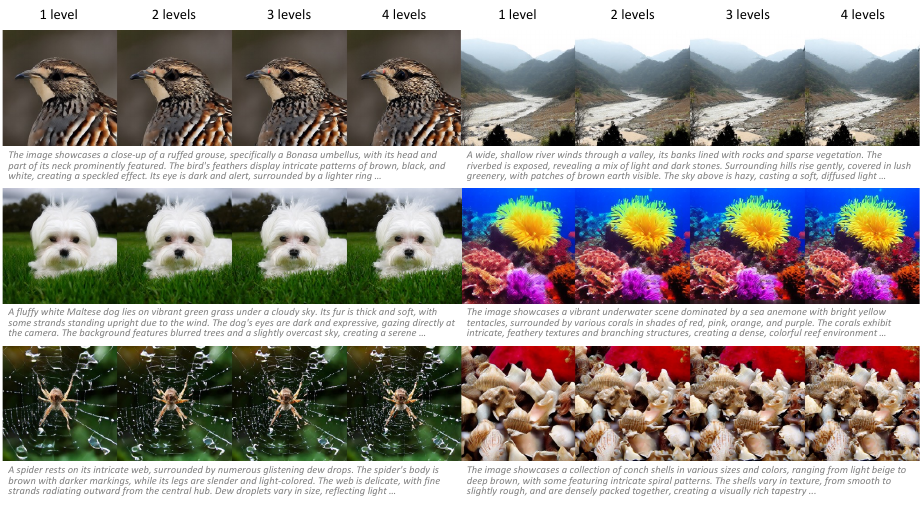}
    \caption{Coarse-to-fine text-to-image generation on ImageNet. Increasing token levels progressively enriches image details. Users can start with fewer token levels for a quick preview and stop once the desired detail is reached.}
    \label{plot:coarse_to_fine}
\end{figure*}

\subsection{Complexity Matters}
\Cref{plot:complexity_recon_image,plot:complexity_recon_video} show the reconstruction quality of different samples at different resolutions and frame rates with various number of latents.
We can draw the following conclusions:

\noindent\textbf{Statistically sublinear scaling between scale and latent capacity.}
For a given sample, to maintain similar reconstruction quality (PSNR), the growth trend of the scale and number of latents exhibits a sublinear relationship.
In contrast, the most widely used linear assumption~\cite{rombach2022ldm,yao2025lightningdit,li2024hunyuan,wan2025wan,yang2024cogvideox} leads to increasingly better reconstruction quality as the scale and latent capacity increase.
On the other hand, fixed-latents strategies similar to VFRTok~\cite{zhong2025vfrtok} result in an inverse relationship between reconstruction quality and scale.
The assumption of DetailFlow~\cite{liu2025detailflow} is relatively reasonable in the statistics of a large number of samples.

\noindent\textbf{Content complexity dominates reconstruction quality.}
Different samples with varying content complexity will result in different rate-distortion curves~\cite{lu2022learning,berger2003rd}.
For example, at the same scale, simple samples can reach reconstruction quality comparable to complex ones using significantly fewer latent tokens.
Similarly, under a fixed latent budget, a large but simple sample may achieve higher reconstruction quality than a small yet complex sample.

Similarly, we can observe the impact of content complexity in generation tasks.
As illustrated in \Cref{plot:complexity_gen}, for relatively simple content, DCS-LDM achieved satisfactory generation results using only tokens of first level.
For complex samples, DCS-LDM used four levels of tokens to render rich details.
This also demonstrates that DCS-LDM provides highly flexible efficiency-quality control by decoupling complexity and scale.

\subsection{Variable Scale Generation}
\noindent\textbf{Resolution \& frame rate.}
As shown in \Cref{plot:var_scale}, DCS-LDM supports generating data at multiple scales via the scale-independent latent space.
Given a latent representation, the DCS-Tok decoder can provide outputs at arbitrary resolutions and frame rates.
For image generation, we fix DCS-DiT to produce four levels of latent tokens and decode them into images ranging from 256px to 512px. (\Cref{plot:var_res})
For video generation, DCS-Tok decode a single level of generated tokens to synthesize results at 12 fps, 24 fps, and 48 fps. (\Cref{plot:var_fps})
Across all scales, the generated outputs preserve consistent semantics and structure.

\noindent\textbf{Aspect ratio.}
\Cref{plot:var_ratio} illustrates the generation results with varying aspect ratios, enabled by the preservation of spatio-temporal structures in the latent space.
For instance, to generate a landscape image with a $h:w=1:2$ aspect ratio, setting the number of patches on the vertical side to $k=16$ yields 32 patches on the horizental side.
This design enables DCS-DiT to infer the desired aspect ratio through noise-shape perception, eliminating the need for extra control signals.

\subsection{Coarse-to-fine Generation}
DCS-LDM supports level-by-level, coarse-to-fine generation, prioritizing the synthesis of essential structures before progressively adding finer details.
\Cref{plot:coarse_to_fine} shows samples generated using one to four token levels. 
For each row, adding more tokens enhances the visual details of the sample.
This paradigm allows users to quickly generate draft ideas at lower computational cost and latency, progressively refining them by adding more tokens until the desired quality is reached.

\definecolor{backblue}{RGB}{232,238,252}
\begin{table*}[t]
    \centering
    \small
    \caption{System-level comparison on ImageNet $256\times256$ label-to-image task, including autoregressive and latent diffusion approaches. DCS-LDM can incorporate semantic alignment techniques, as in REPA, LightningDiT, and MAETok, to further improve generation quality.}
    \setlength{\tabcolsep}{3.5pt}
    \begin{tabular}{lccccccccccc}
    \toprule
    \multirow{2}{*}{\textbf{Method}} & \multirow{2}{*}{\textbf{\#Params}} & \multicolumn{6}{c}{\textbf{Reconstruction}} & \multicolumn{2}{c}{\textbf{w/o CFG}} & \multicolumn{2}{c}{\textbf{w/ CFG}} \\
    \cmidrule(lr){3-8} \cmidrule(lr){9-10} \cmidrule(lr){11-12}
     &  & \textbf{Tokenizer} & \textbf{\#Token} & \textbf{\#Dim} & \textbf{PSNR}$\uparrow$ & \textbf{SSIM}$\uparrow$ & \textbf{rFID}$\downarrow$ & \textbf{gFID}$\downarrow$ & \textbf{IS}$\uparrow$ & \textbf{gFID}$\downarrow$ & \textbf{IS}$\uparrow$ \\
    \midrule
    MaskGiT~\cite{chang2022maskgit} & 227M & MaskGiT & 256 & 256 & - & - & 2.28 & 6.18 & 182.1 & - & - \\
    LlamaGen~\cite{sun2024llamagen} & 3.1B & VQGAN~\cite{esser2021vqgan} & 256 & 8 & 20.79 & 0.675 & 2.19 & 9.38 & 112.9 & 2.18 & 263.3 \\
    FlexTok~\cite{bachmann2025flextok} & 1.3B & FlexTok & 256 & 6 & - & - & 1.45 & 2.50 & - & 1.86 & - \\
    DetailFlow~\cite{liu2025detailflow} & 326M & DetailFlow & 256 & 8 & 20.08 & 0.670 & 0.80 & - & - & 2.62 & - \\
    \midrule
    DiT~\cite{peebles2023dit} & \multirow{6}{*}{675M} & \multirow{3}{*}{SD-VAE~\cite{rombach2022ldm}} & \multirow{3}{*}{1024} & \multirow{3}{*}{4} & \multirow{3}{*}{26.04} & \multirow{3}{*}{0.834} & \multirow{3}{*}{0.62} & 9.62 & 121.5 & 2.77 & 278.2 \\
    SiT~\cite{ma2024sit} & & & & & & & & 8.61 & 131.7 & 2.06 & 270.3 \\
    REPA~\cite{yu2024repa} & & & & & & & & 5.90 & 157.8 & 1.42 & 305.7 \\
    LightningDiT~\cite{yao2025lightningdit} & & VA-VAE & 256 & 32 & 26.30 & 0.846 & 0.28 & 2.17 & 205.6 & 1.35 & 295.3 \\
    MAETok~\cite{chen2025maetok} & & MAETok & 128 & 32 & 23.61 & 0.763 & 0.48 & 2.56 & 224.5 & 1.72 & 307.3 \\
    \rowcolor{backblue} Ours & & DCS-Tok & 256 & 16 & 26.91 & 0.778 & 0.90 & 7.68 & 111.4 & 2.50 & 281.8 \\
    \bottomrule
    \end{tabular}
    \label{tab:image_compare}
\end{table*}
\begin{table}[t]
    \centering
    \small
    \caption{System-level comparison on UCF101 24fps and $256\times256$ label-to-video task.}
    \setlength{\tabcolsep}{2pt}
    \begin{tabular}{lcccccc}
    \toprule
    \multirow{2}{*}{\textbf{Method}} & \multicolumn{4}{c}{\textbf{Reconstruction}} & \textbf{w/o CFG} & \textbf{w/ CFG} \\
    \cmidrule(lr){2-5} \cmidrule(lr){6-6} \cmidrule(lr){7-7}
     & \textbf{\#Tok} & \textbf{\#Dim} & \textbf{PSNR}$\uparrow$ & \textbf{rFVD}$\downarrow$ & \textbf{gFVD}$\downarrow$ & \textbf{gFVD}$\downarrow$ \\
    \midrule
    Omni.~\cite{wang2024omnitokenizer} & 4096 & 8 & 28.95 & 10.21 & 480.60 & 88.89 \\
    Cosmos~\cite{agarwal2025cosmos} & 2048 & 16 & 31.70 & 13.67 & 497.01 & 85.22 \\
    VFRTok~\cite{zhong2025vfrtok} & 512 & 32 & 31.54 & 13.79 & 377.50 & 71.34 \\
    \rowcolor{backblue} Ours & 1024 & 16 & 28.55 & 17.51 & 452.99 & 86.87 \\
    \bottomrule
    \end{tabular}
    \label{tab:video_compare}
\end{table}

\subsection{System-level Comparison}\label{plot:compare}
We also compare DCS-LDM with existing image and video generation frameworks~\cite{liu2025detailflow,peebles2023dit,yao2025lightningdit,zhong2025vfrtok,wang2024omnitokenizer}.
To ensure fairness, we follow the experimental settings commonly adopted in prior work~\cite{yao2025lightningdit,chen2025maetok,zhong2025vfrtok}, evaluating image reconstruction and label-to-image generation at a resolution of $256\times256$ on ImageNet~\cite{deng2009imagenet}, and video reconstruction and label-to-video generation at $256\times256$ and 24 fps on UCF101~\cite{soomro2012ucf101}.
Across all tasks, DCS-LDM only use the first level tokens.
DCS-LDM delivers reconstruction and generation quality on par with vanilla LDMs~\cite{song2020ddim,ma2024sit,wang2024omnitokenizer,agarwal2025cosmos,zhong2025vfrtok}, as detailed in \Cref{tab:image_compare,tab:video_compare}.
Furthermore, incorporating techniques such as representation alignment in RePA~\cite{yu2024repa}, LightningDiT~\cite{yao2025lightningdit}, and MAETok~\cite{chen2025maetok} into DCS-LDM may further enhance the quality of the generation.
These results demonstrate that DCS-LDM’s novel structural design not only introduces flexibility but also maintains competitive performance in both reconstruction and generation.

\begin{figure}[t]
    \centering
    \includegraphics[width=\linewidth]{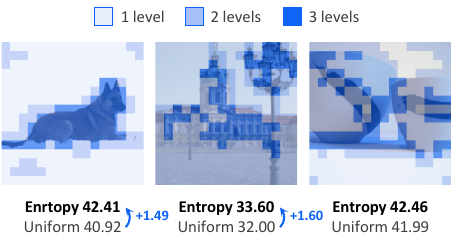}
    \caption{Entropy-guided content-adaptive token allocation. Reconstruction PSNR is measured under an average of two tokens per patch, with a minimum of one and a maximum of three.}
    \label{plot:adaptive}
\end{figure}
\subsection{Content-adaptive Token Allocation}\label{sec:alloc}
DCS-LDM captures the information inhomogeneity among samples by assigning different token levels to different instances.
Moreover, the information distribution within a single sample is often non-uniform~\cite{zhang2025gpstoken}.
For instance, in the example shown in \Cref{plot:simple1}, the foreground hand contains most of the visual information, whereas the background remains relatively simple.
This observation suggests that leveraging such inner-sample non-uniformity can further enhance the efficiency of visual representation.

Since DCS-Tok allocates token levels independently for each patch during training (see \Cref{supp_sec:train}), it naturally supports modeling this non-uniform distribution.
As illustrated in \Cref{plot:adaptive}, we evaluate this property on an image reconstruction task.
Specifically, we compute the entropy of each patch and allocate more tokens to patches with higher entropy.
Under a fixed average token budget of two levels, uniform allocation achieves a PSNR of 34.25, whereas entropy-based allocation attains 34.99 (+0.74) on the ImageNet~\cite{deng2009imagenet} validation set.
Further details of the entropy-based allocation algorithm are provided in \Cref{supp_sec:alloc}.

\noindent\textbf{Future work.}
While DCS-LDM effectively models non-uniform information distribution within samples in reconstruction tasks, further investigation is required to automatically determine the optimal number of tokens for different patches during generation.

%% file: sec/5_conclusion.tex
\section{Conclusion}
We explore the independence between information complexity and data scale in images and videos.
To this end, we propose DCS-LDM, a latent diffusion model that uses variable-level latents to represent information complexity while enabling decoding at arbitrary scales.
This design introduces flexible computation–quality tradeoffs into the generation process.
In addition, DCS-LDM supports a coarse-to-fine progressive generation paradigm, which delivers results more efficiently with the same total computational cost and allows generation to terminate early as needed.
Under this paradigm, DCS-LDM enables rapid semantic and structural previewing of generated samples, followed by fine-grained refinement.
Extensive experiments demonstrate that DCS-LDM achieves performance comparable to state-of-the-art models while offering substantially greater flexibility and functionality.

%% file: sec/X_suppl.tex
\clearpage
\setcounter{page}{1}
\maketitlesupplementary

\section{Implementation Details}\label{supp_sec:details}
\begin{figure}[t]
    \centering
    \begin{subfigure}[t]{0.51\linewidth}
        \centering
        \includegraphics[width=\linewidth]{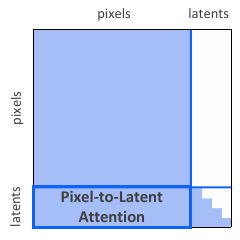}
        \caption{PLA}
        \label{supp_plot:pla}
    \end{subfigure}
    \hfill
    \begin{subfigure}[t]{0.47\linewidth}
        \centering
        \includegraphics[width=\linewidth]{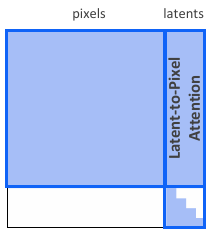}
        \caption{LPA}
        \label{supp_plot:lpa}
    \end{subfigure}
    \\
    \begin{subfigure}[t]{0.51\linewidth}
        \centering
        \includegraphics[width=\linewidth]{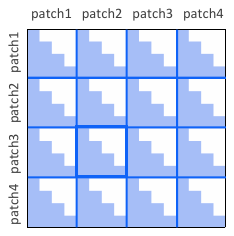}
        \caption{LCA}
        \label{supp_plot:lca}
    \end{subfigure}
    \hfill
    \begin{subfigure}[t]{0.47\linewidth}
        \centering
        \includegraphics[width=\linewidth]{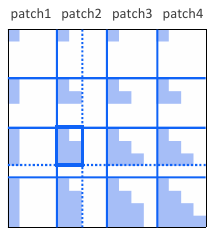}
        \caption{LCA (token drop)}
        \label{supp_plot:lca_patch}
    \end{subfigure}
    \\
    \begin{subfigure}[t]{\linewidth}
        \centering
        \includegraphics[width=0.92\linewidth]{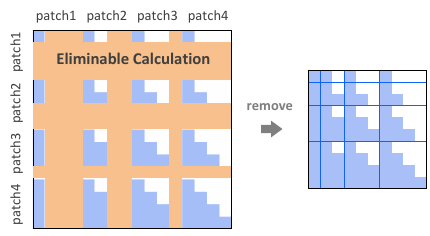}
        \caption{During inference (batch size = 1), unused tokens can be removed.}
        \label{supp_plot:infer_attn}
    \end{subfigure}
    \caption{Attention masks for an image partitioned into $2\times2$ patches and represented with four latent levels.}
    \label{supp_plot:attn_mask}
\end{figure}
\begin{figure}
    \centering
    \includegraphics[width=0.85\linewidth]{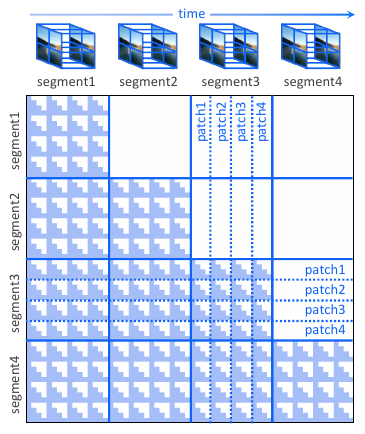}
    \caption{Attention mask for a video partitioned into $4\times2\times2$ patches, and represented with four latent levels.}
    \label{supp_plot:video_attn}
\end{figure}
\subsection{Attention Mask}\label{supp_sec:attn_mask}
As shown in \Cref{fig:arch}, information flow in the patchify layers is directional (PLA and LPA), while tokens in the autoencoder, and DiT follow level-wise causality (LCA).
Consider an example where an image is divided into $2\times2$ patches and encoded using four latent levels. We use attention masks to enforce the intended information flow, as illustrated in \Cref{supp_plot:pla,supp_plot:lpa,supp_plot:lca}.

Since DCS-Tok randomly discards tail tokens during training to structure the latent space, we additionally mask tokens that should not participate in computation.
For instance, \Cref{supp_plot:lca_patch} shows a case where the $i$-th patch is represented using $i$ tokens.
The valid attention regions (in blue) in the third row, second column highlight how the three tokens of the third patch attend to the two tokens of the second patch.
For computational efficiency, we take the maximum number of active tokens across all patches in a micro-batch and remove unused tail levels.

\Cref{supp_plot:lca_patch} contains entire rows and columns that are fully masked—these computations can be eliminated.
With batch size = 1 during inference, the unused tokens for each patch can be directly discarded, as illustrated in \Cref{supp_plot:infer_attn}, leading to more efficient computation.
This optimization is meaningful primarily when token levels vary across patches (\Cref{sec:alloc}); in most settings in this paper, all patches share the same number of levels.

Finally, in video scenarios, we further introduce temporal causality.
As shown in \Cref{supp_plot:video_attn}, temporal causality is applied across segments, where each segment consists of $p_t$ consecutive frames.

\begin{figure}[t]
    \centering
    \begin{subfigure}[t]{\linewidth}
        \centering
        \includegraphics[width=0.9\linewidth]{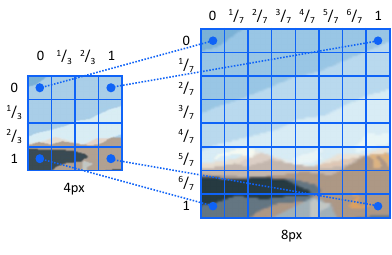}
        \caption{Spatial dimensions are corner-aligned across different resolutions.}
        \label{supp_plot:pos_emb_image}
    \end{subfigure}
    \\
    \begin{subfigure}[t]{\linewidth}
        \centering
        \includegraphics[width=0.6\linewidth]{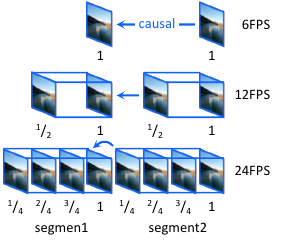}
        \caption{Temporal dimension is tail-aligned across different frame rates.}
        \label{supp_plot:pos_emb_video}
    \end{subfigure}
    \caption{Pixel position representation in the patchifier.}
    \label{supp_plot:pos_emb}
\end{figure}
\subsection{Position Embedding}\label{supp_sec:pos_embed}
DCS-LDM is a pure transformer architecture that employs Rotary Position Embedding (RoPE) to encode positional information.
We partition the channel dimension into multiple groups, assigning them to encode temporal, spatial, and latent-level positions, respectively.
The 1D patchifier and depatchifier operate on positional information within each patch, whereas the autoencoder and DiT define positional information across patches.
For clarity and structural consistency, we use the 1D patchifier and the autoencoder as illustrative examples in the following explanation.

\noindent\textbf{1D patchifier.}
The 1D patchifier contains heterogeneous pixel tokens and latent tokens.
Accordingly, pixel tokens have no latent-level information, while latent tokens do not encode spatio-temporal positions.
As described in \Cref{sec:latents_level}, the 1D patchifier must handle scale-aware patches, therefore pixel position information must be adjusted according to the input scale.
As shown in \Cref{supp_plot:pos_emb}, DCS-LDM normalizes positional coordinates along both spatial and temporal dimensions within each patch, so increases in resolution or frame rate are interpreted as interpolations of positional information.

For the spatial dimensions, we adopt corner alignment: the four corner pixels are always assigned the fixed coordinates (0,0), (0,1), (1,0), and (1,1), and positions for the remaining pixels are obtained via interpolation (\Cref{supp_plot:pos_emb_image}).

For the temporal dimension, illustrated in \Cref{supp_plot:pos_emb_video}, we use a tail-aligned scheme across different frame rates, ensuring that the last frame always has a temporal position of 1.
For images, the temporal dimension is also standardized to 1.
We adopt tail alignment in the temporal dimension because, in asymmetric encoding and decoding, the model may need to reconstruct high–frame-rate video from low–frame-rate input.
Since DCS-LDM defines causality between segments, the representation of a later patch can retrieve information from preceding ones to predict the earlier missing frames, enforcing temporal consistency.

\noindent\textbf{Autoencoder.}
Since the 1D patchifier output is already scale-independent, the autoencoder only needs to account for variations in aspect ratio.
We treat landscape and portrait layouts as extrapolations from square samples.
Concretely, after normalizing the shorter side to 1, we normalize the longer side to $\max(h,w)/\min(h,w)$.

\begin{table}[t]
    \centering
    \small
    \caption{Model parameter configuration.}
    \begin{tabular}{lccc}
    \toprule
    \textbf{Configuration} & \textbf{Patchifier} & \textbf{Autoencoder} & \textbf{DiT} \\
    \midrule
    hidden dimension & 256 & 768 & 1152 \\
    attention heads & 8 & 12 & 16 \\
    layers & 3 & 12 & 28 \\
    output dimension & 256 & 16 & 16 \\
    \bottomrule
    \end{tabular}
    \label{supp_tab:model_conf}
\end{table}

\begin{table}[t]
    \centering
    \small
    \caption{DCS-Tok training configuration, where Img Pre., Img Asym, and Vid Asym correspond to image pretraining, image asymmetric training, and video asymmetric training, respectively.}
    \begin{tabular}{lccc}
    \toprule
    \multirow{2}{*}{\textbf{Configuration}} & \multicolumn{3}{c}{\textbf{Training Stages}} \\
     & \textbf{Img Pre.} & \textbf{Img Asym.} & \textbf{Vid Asym.} \\
    \midrule
    \multicolumn{4}{l}{\textit{\textbf{data}}} \\
    ImageNet & \Checkmark & \Checkmark & \Checkmark \\
    Koala-36M & \XSolidBrush & \Checkmark & \Checkmark \\
    LAVIB & \XSolidBrush & \XSolidBrush & \Checkmark \\
    min resolution & 256px & 256px & 128px \\
    max resolution & 256px & 720px & 720px \\
    min frame rate & - & - & 12FPS \\
    max frame rate & - & - & 60FPS \\
    \midrule
    \multicolumn{4}{l}{\textit{\textbf{latents}}} \\
    \#levels $n$ & 4 & 8 & 8 \\
    alloc $m$ levels prob. & \multicolumn{3}{c}{$p_m=m/\sum_{i=1}^n i,\,m\in[1,n]$} \\
    \midrule
    \multicolumn{4}{l}{\textit{\textbf{loss}}} \\
    perceptual loss & \Checkmark & \Checkmark & \Checkmark \\
    adversarial loss & \XSolidBrush & \XSolidBrush & \Checkmark \\
    denoising task & \XSolidBrush & \XSolidBrush & \Checkmark \\
    latent regulation & \XSolidBrush & \XSolidBrush & \Checkmark \\
    \midrule
    \multicolumn{4}{l}{\textit{\textbf{optimization}}} \\
    \#steps & 100K & 200K & 150K \\
    batch size (256px) & 128 & 256 & 256 \\
    learning rate & 1e-4 & 5e-5 & 5e-5 \\
    optimizer & \multicolumn{3}{c}{AdamW ($\beta_1,\beta_2=0.9,0.95$)} \\
    \bottomrule
    \end{tabular}
    \label{supp_tab:train_tok}
\end{table}

\begin{table}[t]
    \centering
    \small
    \caption{DCS-DiT training and inference configuration.}
    \begin{tabular}{ll}
    \toprule
    \textbf{Configuration} & \textbf{Value} \\
    \midrule
    \multicolumn{2}{l}{\textit{\textbf{data}}} \\
    ImageNet & encoding at 256px \\
    SA-1B & encoding at 512px \\
    \midrule
    \multicolumn{2}{l}{\textit{\textbf{latents}}} \\
    \#levels $n$ & 4 \\
    alloc $m$ levels prob. & $p_m=m/\sum_{i=1}^n i,\,m\in[1,n]$ \\
    \midrule
    \multicolumn{2}{l}{\textit{\textbf{transpose}}} \\
    rectified flow & \Checkmark \\
    logit normal sampling & \Checkmark \\
    velocity direction loss & \Checkmark \\
    \midrule
    \multicolumn{2}{l}{\textit{\textbf{optimization}}} \\
    \#steps & 200K \\
    batch size & 512 \\
    learning rate & 1e-4 \\
    optimizer & AdamW ($\beta_1,\beta_2=0.9,0.95$) \\
    \midrule
    \multicolumn{2}{l}{\textit{\textbf{inference}}} \\
    diffusion sampler & Euler \\
    \#sampling steps & 50 \\
    CFG scale & 6 \\
    CFG interval & 0.1 \\
    timestamp shift & 1 \\
    \bottomrule
    \end{tabular}
    \label{supp_tab:train_dit}
\end{table}

\subsection{Training Details}\label{supp_sec:train}
We first present the model configurations in \Cref{supp_tab:model_conf}.
Training details are provided in \Cref{supp_tab:train_tok,supp_tab:train_dit}, respectively.

\noindent\textbf{DCS-Tok} is trained in multiple stages, including image pretraining, image asymmetric training, and video asymmetric training.
In the image pretraining stage, we use only ImageNet~\cite{deng2009imagenet} to quickly initialize the model with smaller batch sizes (128) and fewer latent levels (4) at a resolution of 256px.
In the image asymmetric training stage, we introduce the Koala-36M~\cite{wang2025koala} dataset and extract only the first frame for training. 
Images range from 256px to 720px, and we enforce corner-aligned scaling (\Cref{supp_sec:pos_embed}) by setting \texttt{align\_corners=True} in the resize operator.
In the final stage, we extend asymmetric encoding to videos with varying frame rates using models trained on the high–frame rate dataset LAVIB~\cite{stergiou2024lavib}. Frame extraction also follows the tail-alignment rule in \Cref{supp_sec:pos_embed}.

During video asymmetric training, we additionally include adversarial loss, a denoising task~\cite{yang2025denoise}, and latent-space regularization. 
During the final 50K steps, we freeze the encoder and fine-tune the decoder to restore reconstruction performance, without applying any denoising tasks~\cite{chen2025maetok}.
Intuitively, the denoising task enhances generation performance, while the margin-based L2 regularization constrains the latent value range and preserves reconstruction quality.
Further details are provided in \Cref{supp_sec:latent_ablation}.

Across all stages, we discard tail latent tokens to structure the latents. Given a maximum of $n$ levels, the model samples $1 \leq m \leq n$ levels for each patch. 
Uniform sampling biases optimization toward early tokens: first-level tokens are always used, while last-level tokens are selected with probability only $1/n$. 
To mitigate this imbalance, we adopt a linear sampling strategy, as summarized in \Cref{supp_tab:train_tok}. 
After applying this scheme, the contribution of later-level tokens to reconstruction quality improves substantially.

\noindent\textbf{DCS-DiT} is trained on the ImageNet~\cite{deng2009imagenet} and SA-1B~\cite{kirillov2023SAM} datasets, and uses the same linear probability sampling strategy as DCS-Tok, except that all patches within each sample share the same number of levels. 
DCS-DiT follows a transpose configuration similar to LightningDiT~\cite{yao2025lightningdit}.

\begin{algorithm}[tp]
    \small
    \caption{Entropy-guided Token Allocation} 
    \label{alg:token_alloc} 
    \KwIn{Patches $X$, desired avg tokens $n$, bounds $[m,M]$, max adjustment $K$, scaling factors $\theta_1,\theta_2\in(0,1)$}
    \KwOut{Token grid $G$}
    $E \gets \text{GetEntropy}(X)$ \\
    $w \gets \text{Normalize}(E,m,M)$ \\
    $w_{\mathrm{best}} \gets w$ \\
    \tcp{Iteratively adjustment}
    \For{$t = 1$ \KwTo $K$}{
        $\hat{n} \gets \text{Mean}(\text{Round}(w))$ \tcp*[f]{Estimation}
        
        $w \gets n/\hat{n} \cdot w$ \tcp*{Rescale}
        
        $\hat{n}_{\mathrm{new}} \gets \text{Mean}(\text{Round}(w))$ \\
        $\hat{n}_{\mathrm{best}} \gets \text{Mean}(\text{Round}(w_{\mathrm{best}}))$ \\
        \If{$|\hat{n}_{\mathrm{new}} - n| < |\hat{n}_{\mathrm{best}} - n|$}{
            $w_{\mathrm{best}} \gets w$
        }
        \tcp{If token count exceeds target, slightly suppress weights}
        \If{$\hat{n}_{\mathrm{new}} > n$}{
            $w \gets \theta_1 \cdot w$
        }
    }
    \tcp{Ensure not exceeds $n$}
    $G \gets \text{Clamp}(\text{Round}(w_{\mathrm{best}}), m, M)$ \\
    \While{$\text{Mean}(G) > n$}{
        $w_{\mathrm{best}} \gets \theta_2 \cdot w_{\mathrm{best}}$ \\
        $G \gets \text{Clamp}(\text{Round}(w_{\mathrm{best}}), m, M)$ \\
    }

    \textbf{Return} $G$
\end{algorithm}
\subsection{Content-adaptive Token Allocation}\label{supp_sec:alloc}
\Cref{alg:token_alloc} presents the pseudo-code for our entropy-guided content-adaptive token allocation strategy in the reconstruction task.
Using the entropy of each patch as a proxy for visual complexity, we assign more tokens to patches with higher entropy and iteratively refine the allocation to satisfy constraints on the average, minimum, and maximum number of tokens per patch.
In practice, we use hyperparameters $K=10$, $\theta_1=0.995$, and $\theta_2=0.999$.
Although this method is relatively naive, it still yields substantial improvements, as demonstrated in \Cref{sec:alloc}.

\begin{figure*}[t]
    \centering
    \begin{subfigure}[t]{\textwidth}
        \centering
        \includegraphics[width=\textwidth]{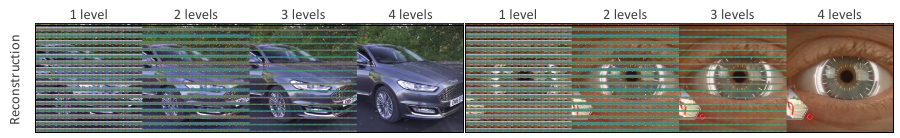}
        \caption{Unstructured latent representation.}
        \label{supp_plot:wo_drop}
    \end{subfigure}
    \\
    \begin{subfigure}[t]{\textwidth}
        \centering
        \includegraphics[width=\textwidth]{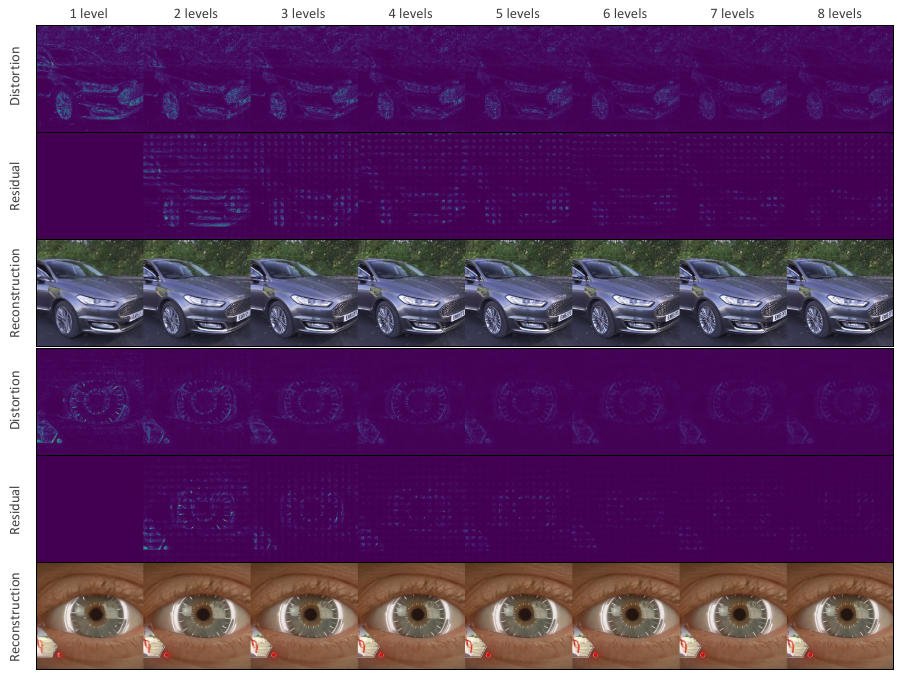}
        \caption{Structured latent representation.}
        \label{supp_plot:w_drop}
    \end{subfigure}
    \caption{Visualization of token encoding information at each level. Distortion represents the difference between the reconstructed and ground-truth images, while residual represents the changes introduced by current level.}
    \label{supp_plot:encode}
\end{figure*}

\section{Ablation Study}
\subsection{Encoded Information of Tokens}
As shown in \Cref{supp_plot:encode}, we visualize reconstruction quality as more latent levels are appended.
\Cref{supp_plot:wo_drop} presents an experimental variant without the token-dropping strategy, where four latent levels are used. 
In this case, the model effectively encodes the image by dividing it into four horizontal bands, indicating that the latent levels do not form a coherent hierarchical representation.
In contrast, \Cref{supp_plot:w_drop} demonstrates that our structured representation captures global image information starting from the first level. As additional levels are introduced, fine details are progressively refined. 
Notably, the improvement is concentrated in high-complexity regions, whereas low-complexity regions are well reconstructed using fewer levels.
This observation directly motivated our exploration of content-adaptive token allocation (\Cref{sec:alloc}).

\definecolor{darkgreen}{RGB}{1,150,32}
\definecolor{darkred}{RGB}{150,32,1}
\begin{table*}[t]
    \centering
    \small
    \caption{Generation quality, latent standard deviation per level (lvl-$i$), and reconstruction quality with varying numbers of latent levels ($j$ lvls) under different training strategies. All implementations that use denoising tasks also include the decoder fine-tuning stage. \textit{Marg.}, \textit{Gen.}, and \textit{Frag.} represents Margin, Generation, and Fragmentation, respectively.}
    \setlength{\tabcolsep}{2.2pt}
    \begin{tabular}{cccccccccccccccccccc}
    \toprule
    \multicolumn{3}{c}{\textbf{Methods}} & \textbf{Gen.} & \multicolumn{8}{c}{\textbf{Latents Standard Deviation}} & \multicolumn{8}{c}{\textbf{Reconstruction PSNR}$\uparrow$} \\
    \cmidrule(lr){1-3} \cmidrule(lr){4-4} \cmidrule(lr){5-12} \cmidrule(lr){13-20}
    \textbf{Denoise} & \textbf{L2} & \textbf{Marg.} & \textbf{Frag.} & \textbf{lvl-1} & \textbf{lvl-2} & \textbf{lvl-3} & \textbf{lvl-4} & \textbf{lvl-5} & \textbf{lvl-6} & \textbf{lvl-7} & \textbf{lvl-8} & \textbf{1 lvl} & \textbf{2 lvls} & \textbf{3 lvls} & \textbf{4 lvls} & \textbf{5 lvls} & \textbf{6 lvls} & \textbf{7 lvls} & \textbf{8 lvls} \\
    \midrule
    \XSolidBrush & \XSolidBrush & - & \color{darkred}\Checkmark & 3.64 & 3.45 & 3.41 & 3.42 & 3.31 & 3.09 & 2.72 & 2.75 & 24.22 & 26.76 & 28.58 & 29.98 & 31.27 & 31.98 & 32.72 & 33.22 \\
    \Checkmark & \XSolidBrush & - & \color{darkred}\Checkmark & \color{darkred}{82.00} & \color{darkred}{67.50} & \color{darkred}{61.25} & \color{darkred}{49.75} & \color{darkred}{50.10} & \color{darkred}{39.99} & \color{darkred}{43.24} & \color{darkred}{40.68} & 24.03 & 26.74 & 28.33 & 29.77 & 31.26 & 31.79 & 32.72 & 33.15 \\
    \Checkmark & \Checkmark & 0 & \color{darkgreen}\XSolidBrush & 12.56 & 6.44 & 4.91 & 2.58 & 1.14 & 0.75 & 0.39 & 0.45 & 24.12 & 26.50 & 28.16 & 29.50 & 29.63 & \color{darkred}{29.64} & \color{darkred}{29.63} & \color{darkred}{29.64} \\
    \Checkmark & \Checkmark & 1 &  \color{darkgreen}\XSolidBrush & 12.44 & 7.38 & 6.09 & 4.97 & 4.28 & 3.61 & 3.03 & 2.69 & 24.07 & 26.44 & 28.17 & 29.48 & 30.71 & 31.35 & 32.09 & 32.61 \\
    \bottomrule
    \end{tabular}
    \label{supp_tab:denoise}
\end{table*}

\begin{figure}[t]
    \centering
    \includegraphics[width=\linewidth]{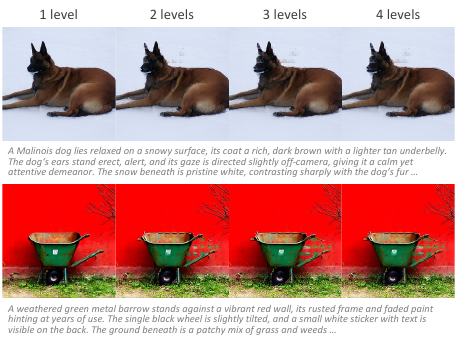}
    \caption{Generation results from a naive implementation without latent denoising and regularization. High-level tokens are harder to model, leading to fragmented generations.}
    \label{supp_plot:broken}
\end{figure}

\subsection{Latent Denoising and Regularization}\label{supp_sec:latent_ablation}
As shown in \Cref{supp_plot:broken}, a naive implementation with poor diffusibility produces fragmented outputs. 
As token levels increase, details are added incorrectly, degrading overall quality.

To address this issue, we introduce the denoising task~\cite{yang2025denoise} to enhance the diffusibility of latent representations. 
During training, we perturb the latent variables with Gaussian noise, as formulated in \Cref{supp_eq:denoise}.
\begin{equation}
\hat z = \lambda\cdot z+(1-\lambda)\cdot \epsilon,\,\lambda\in \mathcal{U}(0,1),\,\epsilon\in \mathcal{N}(0,\sigma),
\label{supp_eq:denoise}
\end{equation}
where we follow prior work~\cite{yang2025denoise} and set $\sigma = 3$.
Meanwhile, we introduce an L2 regularization with margin to constrain the numerical range of latents without hindering high-level token reconstruction. 

As shown in \Cref{supp_tab:denoise}, we compare generation and reconstruction quality across different configurations. 
We find that applying the denoising task alone is ineffective. 
Because DCS-Tok is a deterministic autoencoder without a KL constraint, its latent values tend to expand rapidly during training to counteract Gaussian noise perturbations, ultimately preventing improvements in diffusibility. 
To address this, we add the L2 regularization to suppress this expansion. 
With this modification, the fragmentation issue is significantly alleviated.

However, we further observed that tokens in higher levels contributed almost nothing to reconstruction, and adding more levels failed to improve quality. 
The numerical range of these higher-level representations was also significantly smaller than that of lower ones. 
Since higher levels have a weaker influence on reconstruction, the model tends to suppress them to minimize the L2 loss.
To counter this effect, we introduce a margin in the L2 regularization to preserve the numerical range of higher levels. 
Experiments show that this modification effectively restores the usefulness of higher-level tokens while eliminating fragmentation.

It is worth noting that, for a fair comparison with existing approaches~\cite{peebles2023dit,ma2024sit}, the model in \Cref{plot:compare} was trained without the denoising task.
